\documentclass[conference]{IEEEtran}
\usepackage{cite}
\ifCLASSINFOpdf
\else
\fi

\usepackage{graphicx}
\usepackage{hyperref}

\begin{document}
\title{GENEVA: GENErating and Visualizing branching narratives using LLMs}

\author{\IEEEauthorblockN{Jorge Leandro, Sudha Rao, Michael Xu, Weijia Xu,}
\IEEEauthorblockN{Nebojsa Jojic, Chris Brockett, Bill Dolan}
\IEEEauthorblockN{Microsoft Research}
\IEEEauthorblockN{Email: Sudha.Rao@microsoft.com}
}



%Telephone: (800) 555--1212, Fax: (888) 555--1212}


\maketitle

\begin{abstract}
    Dialogue-based Role Playing Games (RPGs) require powerful storytelling. The narratives of these may take years to write and typically involve a large creative team. In this work, we demonstrate the potential of large generative text models to assist this process. \textbf{GENEVA}, a prototype tool, generates a rich narrative graph with branching and reconverging storylines that match a high-level narrative description and constraints provided by the designer. A large language model (LLM), GPT-4, is used to generate the branching narrative and to render it in a graph format in a two-step process. We illustrate the use of GENEVA in generating new branching narratives for four well-known stories under different contextual constraints. This tool has the potential to assist in game development, simulations, and other applications with game-like properties.
\end{abstract}


% \fi
\IEEEpeerreviewmaketitle

\section{Introduction}

\begin{figure*}[h!t]
    \centering
    \includegraphics[scale=0.5]{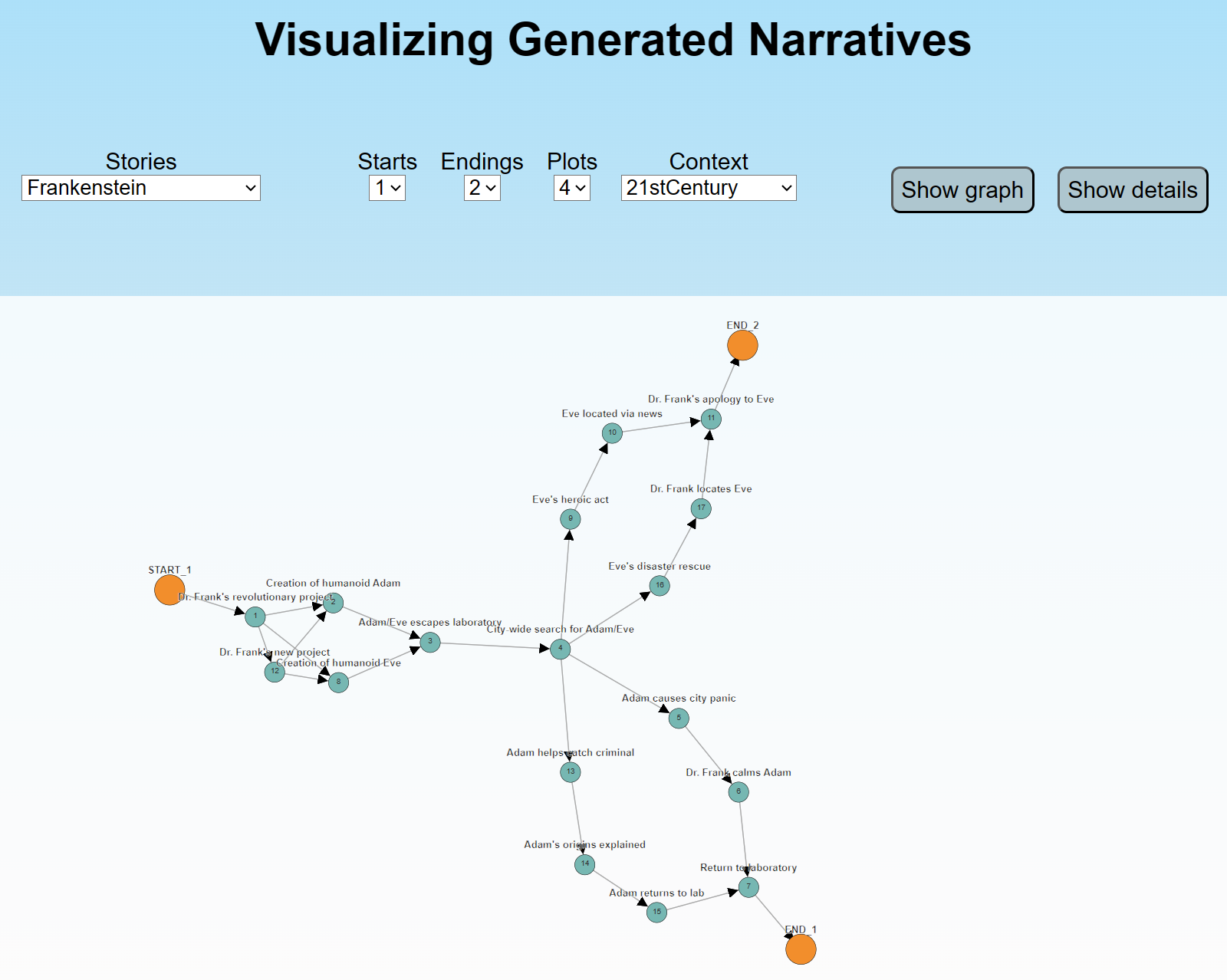}
    \caption{In the online interface of GENEVA, you can choose a particular story, pick the number of starts, ends and storylines and click on the `Show Graph' button to see the narrative graph for that story. The above figure shows the narrative graph for the Frankenstein story but grounded in the 21st century. Additional constraints on the graph includes one start, two endings and four storylines.}
    \label{fig:frankenstien-1-2-4-21stcentury}
\end{figure*}

Whether we are talking about a movie, a novel, or a game, successful storytelling requires a rich narrative structure with an engaging plot and a satisfying climax. Today such design goals are met through heavy manual authoring: dialog trees, plot outlines, and branching narratives. The task can be enormous, and it becomes even more difficult as players seek a vast and potentially limitless world. In the past, there has been work on creating tools to aid game designers \cite{grabska2021application, neil2012game,alvarez2022story,liapis2013designer}. None, however, involve the latest generative LLMs in this creative process. We, by contrast, seek to leverage the inherent confabulatory capacity of the model for creative ends. We present GENEVA, a graph-based narrative generation and visualization tool that uses large generative text models. 

Input to GENEVA comprises of a high-level narrative description and a list of constraints: number of different starts, number of different endings, number of different storylines and a context for grounding the narrative. Given these inputs, it uses a large generative language model, GPT-4 \cite{Koubaa2023}, to both generate a narrative with branching storylines and to render this narrative in a graph format. Figure \ref{fig:frankenstien-1-2-4-21stcentury} shows the narrative graph created by GENEVA for the Frankenstein story set in the 21st century\footnote{The online interface of GENEVA: \url{https://narrative.msr-emergence.com/} and an online demo: \url{https://www.youtube.com/watch?v=SE0SEd6nXXU}}. The narrative graph is essentially a DAG (directed acyclic graph) in which each node represents a ``narrative beat'' and the directed edges represent the temporal progression through the beats. A narrative (or story) beat is ``the smallest element of structure,'' ``an exchange of behavior in action/reaction'' \cite[p. 37]{McKee1997}. A single path from one of the start nodes to one of the end nodes defines a unique storyline and the narrative graph as a whole depicts the different storylines possible with the same higher-level narrative. Figure \ref{fig:frankenstien-1-2-4-21stcentury-details} shows the different storylines in the narrative graph and the sequence of beats that make up each of the storyline.

We implement the generation and the rendering of the narrative graph using Open AI's GPT-4 \cite{Koubaa2023}.
We design two prompts: the first takes the description and constraints as input and outputs the branching storylines in text format; the second prompt takes those storylines and outputs code to render them in a graph format. We conduct a case study where we use GENEVA to create narrative graphs for four well-known stories: Dracula, Frankenstein, Jack and the Beanstalk, and Little Red Riding Hood. These are chosen for their familiarity, so that it is easy to see the scope of the generated variations. The high-level narrative descriptions are simply the titles of the stories. We ground these in one of the four settings: Minecraft game, 21st century, Ancient Rome or Quantum Realm. We also experiment with constraints specific to the graph structure: number of different starts, number of different endings and number of different storylines. We include an analysis of the narrative graphs identifying some of its strengths and weaknesses.

\section{METHODOLOGY}
\label{sec:geneva-implementation}

We use GPT-4 to generate the narrative graph in two steps. First we prompt GPT-4 to create the narrative graph, with branching storylines in text format. Then we use GPT-4 to encode this information into a format that can be consumed by the visualization module. 

\subsection{Generating the Story Lines}
\label{sec:generate-storylines}

We define a storyline as a sequence of narrative beats. Narrative or story beats are often defined as significant moments in a story that evoke particular emotional reaction from the audience. We treat narrative beats as the building blocks of a storyline. Our first step is to create a prompt that includes the high-level narrative description, the input constraints, and specific instructions on generating multiple branching story lines that will make up the narrative graph. By way of example, let's say we want to generate a branching narrative for the well-known Frankenstein story, but have it translated into a 21st century context. Additionally we want the narrative graph to have one starting point, two different endings, and a total of 4 different storylines.

We prompt GPT-4 to follow an iterative process where we ask it to first create a single storyline by generating a sequence of beats and describing each beat in detail. We then instruct it to generate the next storyline such that it contains some beats in common with the first and some new beats and finally keep repeating this process until it generates all the storylines. Additionally, we ask GPT-4 to strictly follow a set of guidelines and constraints sketched below: 
\begin{itemize}
\setlength\itemsep{0.1em}
    \item Each storyline must follow the conventions of a narrative consisting of conflicts or danger with clear resolutions and no loose ends.
    \item Storylines must be uniquely different, no more than three same consecutive beats between any two storylines.
    \item Total number of unique beats must be twice the number of required storylines.
    \item Original story must appear as one storyline.
    \item Ground storyline in unique characteristics of the input, including cultural elements, timeline, technology, etc.
    \item Must have as many unique starts and ends as requested.
    \item Must have 2 or 3 common beats between all storylines.
\end{itemize}

\begin{figure*}[h!t]
    \centering
    \includegraphics[scale=0.38]{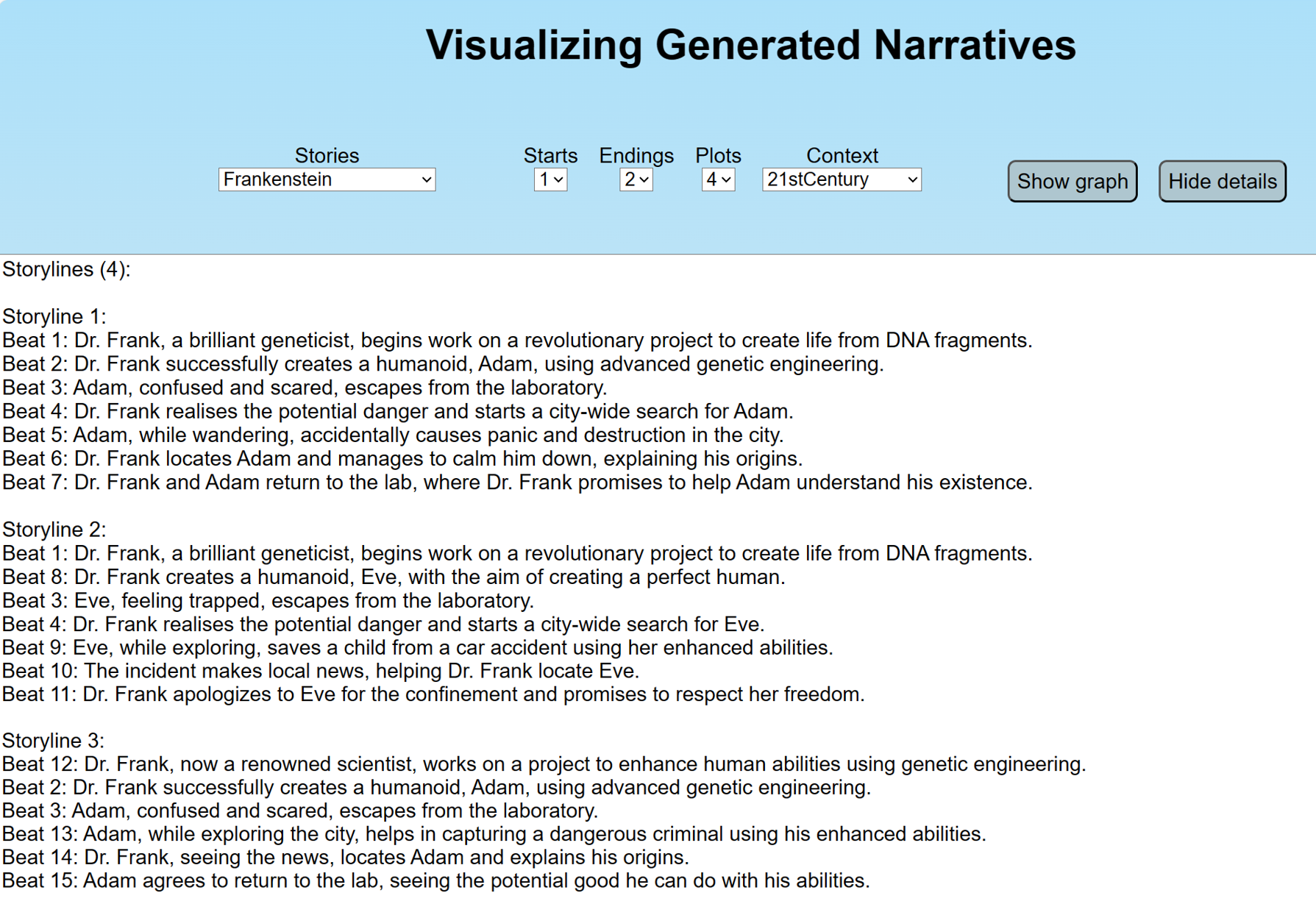}
    \caption{In the online interface of GENEVA, you can see the detailed description of the sequence of beats that make up each storyline by clicking on the `Show Details' button. The above figure shows detailed view of the four different storylines present in the narrative graph in Figure 1. }
    \label{fig:frankenstien-1-2-4-21stcentury-details}
\end{figure*}

Finally, we include an example input and the expected output in the prompt. This prompt\footnote{Full prompts are included in the appendix.} results in the generation of storylines in text format as illustrated in Figure \ref{fig:frankenstien-1-2-4-21stcentury-details}.

Next, we highlight some of our findings during the process of crafting this prompt. The most important one was that when asked to generate all storylines at once, the model was not good at interweaving the storylines to create a true branching narrative. More specifically, it generated independent storylines that did not intersect to create the branching/converging structure. Instead when we replaced this top-down approach with a bottom-up iterative approach where we instructed the model to generate only one storyline first and then generate the next one while keeping the first one in mind, the generated storylines ended up creating a branching structure where the storylines intersected and diverged at crucial points. Another finding was that including an example input-output pair helped model in generating the storylines in a consistent format that is important for us since the output of this step goes as input to our next step (i.e. generating the visualization). 

% \textcolor{magenta}{JORGE ADDS: In terms of prompt shape, prompt design process and prompt strategy\cite{pengfei2021}, our prompts are respectively \emph{prefix} prompts, \emph{manually crafted} prompts and \emph{dynamic}. It is worth mentioning that the dynamic aspect of our prompts are due to the injection of different pieces of data to get a custom prompt for each input. Manually crafted prompts result from an intuitive trial-and-error process, built upon human introspection, therefore subject to drawbacks such as depending on the available time, the designer experience, without guarantee of finding an optimal prompt\cite{pengfei2021}.}

% We highlight some findings with respect to the process of hand-crafting the prompt, but encourage the reader to infer more insightful clues from the provided prompts themselves.
% \begin{itemize}
% \setlength\itemsep{0.1em}
% \item avoid ambiguous instructions at all costs
% \item pay attention to semantics to avoid conflicting instructions
% \item in general, experiments show that examples (few-shot) supersede instructions
% \item sometimes, the undesired behavior must also be stated: "Don't itemize or number outputs"
% \item instructions must explicitly require semantic uniqueness
% \item part-whole behavior: experiments show that the model gets confused if requested to generate all stories from a top-bottom approach, yielding unstable results. A bottom-up iterative approach led to stable results.
% \end{itemize}

\subsection{Generating the Visualization}
\label{sec:generating-visualization}

The next step is to generate a visual graph of the generated storylines. We prompt GPT-4 with the generated storylines and additional instructions on how to generate the graph in a particular convention of nodes and edges, as input data to a Javascript D3JS browser application. We ask GPT-4 to strictly adhere to a set of guidelines sketched as below:
\begin{itemize}
\setlength\itemsep{0.1em}
    \item Create a node for each beat such that the total number of nodes equal total number of beats.
    \item Create an edge between each pair of adjacent nodes in the sequence of storylines.
    \item Every node should be connected to the graph. 
    \item Create a NODES object as a dictionary with key as a beat number and value as the beat description.
    \item Create an EDGES object as a dictionary with key as a node and value as a list incoming and outgoing edges.
    \item Make sure that every node in the NODES object also appears in the EDGES object and vice-versa.
\end{itemize}

% \textcolor{magenta}{JORGE ADDS: The LLM GPT-4 has been also leveraged for the summarization task during the graph generation and rendering, so as to create labels for every node.}

Finally, we include an example of input storylines and output graph in the prompt. This prompt leads to the generation of the final narrative graph illustrated in Figure \ref{fig:frankenstien-1-2-4-21stcentury}. We also use GPT-4 to generate a summary of each beat description and use this summarized beat descriptions as node labels to be displayed on the graph. 

We highlight some of our findings from crafting this prompt. We found that mentioning explicitly the desired properties of the graph is very important. For example, there should be no disconnected nodes, edges should be directed, NODES and EDGES objects should be consistent, and there should be a one-to-one mapping  between the beats and the nodes. Also, as in the previous step, including an example input-output pair helps in getting a consistent output that adheres to the format expected by the Javascript D3JS browser application that renders the graph. 

% As for the graph structure generation, the desired properties of the graph must be explicitly stated:
% \begin{itemize}
%  \item  no disconnected nodes
%  \item edges direction
%  \item location of start and end nodes
%  \item consistency between NODES and EDGES objects
%  \item one-to-one mapping between beats and nodes
% \end{itemize}
\section{Case Study}

\subsection{Stories, settings and constraints}

 We experiment with four well-known stories: Dracula, Frankenstein, Jack and the Beanstalk and Little Red Riding Hood. We consider four settings for grounding these stories: Minecraft, 21st century, Ancient Rome and Quantum Realm. Additionally, we consider the following constraints on the graph structure: Number of starts: [1, 2], Number of endings: [2, 4], Number of storylines: [4, 8]. It is important to note that although we chose these specific stories and settings in the deployed version of GENEVA, the prompts defined in section \ref{sec:geneva-implementation} are generic enough that one can generate such narrative graphs for any story and setting. 

\subsection{Analysis of the generate narrative graphs}

We find that GENEVA is able to ground certain stories quite well in a given setting; while adhering to the constraints on the number of starts, ends and storylines. For example, when asked to ground Little Red Riding Hood in the game of Minecraft, it generates storylines that include healing potions, mob-infested cave, a redstone contraption and other elements that are specific to the game of Minecraft, while still retaining the narrative related to the original little red riding hood story: for example little red riding hood's grandmother asking her to get healing potions, little red encountering a cunning player who manipulates her game environment, and little red exposing the player to the game's moderator. Another interesting observation was that the model chooses to steer clear of the original ending (where little red gets eaten by the wolf) in all the generated storylines and instead chooses to gives them positive endings where little red ends up learning a valuable lesson. 

When asked to ground Frankenstein in the 21st century, the storylines include a project on creating life from DNA fragments, genetic engineering, a humanoid and other aspects that are relatable in the 21st century while still maintaining the theme of the original Frankenstein story. For example, in one storyline, Dr. Frank creates a humanoid Adam, in another he creates Eve. Confused and scared, they escape from the laboratory and there is a city-wide search for them. In one storyline, Eve heroically saves a child and Dr Frank ends up apologizing and promising her freedom. In another storyline, Adam accidentally causes panic and the destruction of the city and Dr. Frank ends up taking him back to the lab and promises to help him understand his existence. As you can see, the language model found interesting variations of the original story to not only ground it in the 21st century but to also make it more palatable to a 21st century audience.  

When asked to ground Dracula in ancient Rome, the storylines include a mysterious sorcerer, a wise oracle, the Colosseum and similar such people, places and objects that are specific to ancient Rome. For example, in one storyline, Dracula is a roman senator who gets turned into a vampire by a sorceror and uses his new powers to control the minds of other senators. A young soldier named Lucius uncovers Dracula's true identity and defeats him in a massive battle in the Roman Forum. We observed that the storylines generated for this combination (i.e. Dracula story in the ancient Rome setting), were not as rich as the ones generated for some of the other combinations. They also tended to deviate much more from the original Dracula story. We attribute this to the fact that the universe in which the original Dracula was set was quite different from ancient Rome and hence the model had difficulty bringing the two together.

Although GENEVA provides a good starting point for a possible narrative, on closer inspection, we find that certain aspects of the generated narratives that can be improved. Firstly, the generated storylines could be more varied. For example, the storylines in Figure \ref{fig:frankenstien-1-2-4-21stcentury} have very similar flavor and could benefit from more variations to make them more interesting. Secondly, in some settings, the grounding could be better. Specifically, grounding in Minecraft or 21st century appears to be much better than grounding in quantum realm. For example, when asked to ground Jack and the Bean Stalk in quantum realm, although the storylines included objects such as the quantum watch, quantum creatures and quantum gems, the plotlines themselves did not adapt well to the quantum realm setting. In one storyline, the only difference from the original was that the beanstalk grows to reach a quantum realm instead of the clouds. Even when the storylines deviated from the original, the changes were much more generic and not specific to the quantum realm. For example, a mysterious old man gives Jack the magic beans instead of Jack trading them for his family cow, or Jack steals a golden harp from the castle instead of gold coins. We presume this is because there is more information about Minecraft and the 21st century in the language model data than there is about quantum realm, information about which may be comparatively limited. This suggests that GENEVA is likely to perform better on better documented settings. Overall, nonetheless, we think that GENEVA is a useful tool for generating a plausible initial narrative that can then be iteratively edited and improved upon by an experienced narrative designer. 
\section{Conclusion}
We have introduced GENEVA, a graph-based narrative visualization that is powered by a large-scale language model. It allows a game designer to input a high-level description of a story with specific constraints and generates a set of branching storylines that can be easily visualized using a graph format.  
We demonstrate its use to generate narrative graphs for some well-known stories by grounding them in interesting alternative settings. We believe that a tool such as this can be a valuable assist to narrative designers allowing them to expend their creativity on defining the higher-level narrative and iterating on it and delegating the lower-level task of generating detailed possible storylines to generative AI models for inspection, approval, and adaptation.

\bibliographystyle{IEEEtran}


% \end{thebibliography}

\bibliography{custom}

\begin{thebibliography}{1}
\providecommand{\url}[1]{#1}
\csname url@samestyle\endcsname
\providecommand{\newblock}{\relax}
\providecommand{\bibinfo}[2]{#2}
\providecommand{\BIBentrySTDinterwordspacing}{\spaceskip=0pt\relax}
\providecommand{\BIBentryALTinterwordstretchfactor}{4}
\providecommand{\BIBentryALTinterwordspacing}{\spaceskip=\fontdimen2\font plus
\BIBentryALTinterwordstretchfactor\fontdimen3\font minus \fontdimen4\font\relax}
\providecommand{\BIBforeignlanguage}[2]{{%
\expandafter\ifx\csname l@#1\endcsname\relax
\typeout{** WARNING: IEEEtran.bst: No hyphenation pattern has been}%
\typeout{** loaded for the language `#1'. Using the pattern for}%
\typeout{** the default language instead.}%
\else
\language=\csname l@#1\endcsname
\fi
#2}}
\providecommand{\BIBdecl}{\relax}
\BIBdecl

\bibitem{grabska2021application}
I.~Grabska-Gradzi{\'n}ska, L.~Nowak, W.~Palacz, and E.~Grabska, ``Application of graphs for story generation in video games,'' in \emph{Proceedings of the 2021 Australasian Computer Science Week Multiconference}, 2021, pp. 1--6.

\bibitem{neil2012game}
K.~Neil, ``Game design tools: Time to evaluate,'' \emph{Proceedings of 2012 DiGRA Nordic}, 2012.

\bibitem{alvarez2022story}
A.~Alvarez, J.~Font, and J.~Togelius, ``Story designer: Towards a mixed-initiative tool to create narrative structures,'' in \emph{Proceedings of the 17th International Conference on the Foundations of Digital Games}, 2022, pp. 1--9.

\bibitem{liapis2013designer}
A.~Liapis, G.~Yannakakis, and J.~Togelius, ``Designer modeling for personalized game content creation tools,'' in \emph{Proceedings of the AAAI Conference on Artificial Intelligence and Interactive Digital Entertainment}, vol.~9, no.~2, 2013, pp. 11--16.

\bibitem{Koubaa2023}
\BIBentryALTinterwordspacing
A.~Koubaa, ``{GPT-4 vs. GPT-3.5: A Concise Showdown},'' 4 2023. [Online]. Available: \url{https://www.techrxiv.org/articles/preprint/GPT-4_vs_GPT-3_5_A_Concise_Showdown/22312330}
\BIBentrySTDinterwordspacing

\bibitem{McKee1997}
R.~McKee, \emph{{Story: Substance, Structure, Style and the Principles of Screenwriting}}.\hskip 1em plus 0.5em minus 0.4em\relax New York: Harper Collins, 1997.

\end{thebibliography}
\section{Appendix}

\subsection{Prompt for generating the story lines}

Here we share the exact prompt we use to generate the storylines (Section \ref{sec:generate-storylines}). 

INSTRUCTION: Your task is to generate unique and interesting storylines given the following INPUT OPTIONS: [include the input story, setting, number of starts, ends and storylines here]\\
Follow the format in the example below, without duplicating its content.\\
Story: (name of the story), \\
Starts: (number of starts here), \\
Endings: (number of endings here), \\
Storylines: (number of storylines here), \\
Setting: (topic on which storylines must be grounded)
    
Storylines (detailed with beat descriptions): \\
Storyline 1: (Line separated sequence of beats. Include a detailed description of each beat and assign it a beat number.) \\
Storyline 2: (Line separated sequence of beats that have some beats common with the previous storyline(s) and some new beats. Include a detailed description of each beat. If the beat is common to one of the previous storylines, then its description and number should be exactly the same as in the previous one as well, but repeat the detailed beat description for clarity. Assign new beat numbers to the new beats.)     \\
… \\
Storyline 10: (Line separated sequence of beats that have some beats common with the previous storyline(s) and some new beats. Include a detailed description of each beat. If the beat is common to one of the previous storylines, then its description and number should be exactly the same as in the previous one as well, but repeat the detailed beat description for clarity. Assign new beat numbers to the new beats)

(List as many dummy start nodes as number of starts in INPUT OPTIONS) \\
START\_1: (This is a dummy node. No description for it. It will always point to the beginning beat of the respective storyline)  \\
START\_2: (This is a dummy node. No description for it. It will always point to the beginning beat of the respective storyline) 

(List as many dummy end nodes as number of starts in INPUT OPTIONS) \\
END\_1: (This is a dummy node. No description for it. The final beat of the respective storyline will point to it) \\
 END\_2: (This is a dummy node. No description for it. The final beat of the respective storyline will point to it) \\
 …

Beats (include the list of all the unique beats from the storylines above. Include the exact same description and exact same beat number) \\
Beat\_1: (beat description) \\
Beat\_2: (beat description) \\
 …   \\
Beat\_n: (beat description)

Common intermediate Beats: (beats numbers that are common to ALL the storylines)\\
    Storylines (with only beat numbers) \\
Storyline 1: (a dummy START node, comma-separated exact sequence of beat numbers of this storyline, a dummy END node) \\
Storyline 2: (a dummy START node, comma-separated exact sequence of beat numbers of this storyline, a dummy END node) \\
 … \\
 Storyline 10: (a dummy START node, comma-separated exact sequence of beat numbers of this storyline, a dummy END node)
 
YOU MUST STRICTLY FOLLOW THESE CONSTRAINTS
    \begin{enumerate}
    \setlength\itemsep{0.1em}
    \item Each storyline must consist of a sequence of narrative beats. Different storylines must have different sequence of beats. The common subsequence between two storylines cannot be greater than three.
    \item THE TOTAL NUMBER OF BEATS MUST BE ATLEAST TWICE THE NUMBER OF STORYLINES. Describe each beat in detail.
    \item Make sure that the original story appears as one of the resulting storylines.  
    \item Ground the storylines in the setting focusing on characteristics of the setting that are unique and help make the storylines interesting and novel. Those characteristics might include cultural elements like foods or clothing or music, strange physical properties, unique flora and fauna, unusual geographical features, and surprising technology.
    \item There must be only as many unique starts as given in the INPUT OPTIONS, with each start pointing to a different beat.
    \item There must be only as many unique endings as given in the INPUT OPTIONS, with each ending being pointed to by a different beat.
    \item THERE MUST BE 2 OR 3 BEATS THAT ARE COMMON IN ALL THE STORYLINES. These must be the important narrative beats in the story. The common beats must not be consecutive.  
    \item IMPORTANT: As you are writing each storyline, think if the sequence of beats make sense to be a coherent storyline. Each storyline should follow the conventions of fairytale narratives of conflicts or dangers and clear resolutions. There should be no loose ends. Each storyline should be a unique sequence of beats that is different from other storylines. 
    \end{enumerate}
    Below is an example output: \\
Story: Little Red Riding Hood \\
Starts: 2 \\
Endings: 4 \\
Storylines: 8 \\
Setting: 21st century

Storylines (8):\\
    Storyline 1: \\
 Beat 1: Red, a tech-savvy girl living in a smart city, receives a call from her sick grandmother. \\
Beat 2: Grandmother requests Red to bring her some medicines from the nearby pharmacy. \\
 Beat 3: Red, wearing her red hoodie, ventures out with her electric scooter. \\
 Beat 4: En route, Red encounters a stranger, a cunning hacker, who learns about her mission. \\
 Beat 5: The hacker manipulates the city's GPS system to mislead Red. \\
 Beat 6: Misled, Red ends up in an abandoned factory. \\
 Beat 7: Realizing the trick, Red uses her tech skills to trace the hacker's location. \\
 Beat 8: Red exposes the hacker to the city's cyber police and continues her journey to her grandmother's house. \\
 Beat 9: Red delivers the medicines and they have a virtual family gathering via video call.

Storyline 2: \\
Beat 1: Red, a tech-savvy girl living in a smart city, receives a call from her sick grandmother. \\
 Beat 10: Grandmother asks Red to bring her a special gadget from the tech mall. \\
 Beat 3: Red, wearing her red hoodie, ventures out with her electric scooter. \\
 Beat 4: En route, Red encounters a stranger, a cunning hacker, who learns about her mission. \\
 Beat 11: The hacker hacks into Red's smartwatch, stealing her personal data. \\
 Beat 12: Red notices suspicious activity on her smartwatch and seeks help from her friend, a cybersecurity expert. \\
 Beat 13: Together, they trace the hacker and retrieve Red's data. \\
 Beat 14: Red buys the gadget and delivers it to her grandmother.

Storyline 3: \\
 Beat 15: Red, a social media influencer, plans a live stream to visit her grandmother. \\
 Beat 2: Grandmother requests Red to bring her some medicines from the nearby pharmacy. \\
 Beat 3: Red, wearing her red hoodie, ventures out with her electric scooter. \\
 Beat 16: Red's live stream attracts the attention of a cyber-stalker. \\
 Beat 17: The stalker tries to find Red's location using the live stream data. \\
 Beat 7: Realizing the threat, Red uses her tech skills to trace the stalker's location. \\
 Beat 8: Red exposes the stalker to the city's cyber police and continues her journey to her grandmother's house. \\
 Beat 9: Red delivers the medicines and they have a virtual family gathering via video call.

Storyline 4: \\
 Beat 15: Red, a social media influencer, plans a live stream to visit her grandmother. \\
 Beat 10: Grandmother asks Red to bring her a special gadget from the tech mall. \\
 Beat 3: Red, wearing her red hoodie, ventures out with her electric scooter. \\
 Beat 16: Red's live stream attracts the attention of a cyber-stalker. \\
 Beat 18: The stalker tries to manipulate Red's followers against her. \\
Beat 19: Red, noticing the unusual comments, uses her influence to expose the stalker's intentions. \\
 Beat 20: Red's followers, united, report the stalker leading to his arrest. \\
Beat 14: Red buys the gadget and delivers it to her grandmother.

Storyline 5: \\
 Beat 1: Red, a tech-savvy girl living in a smart city, receives a call from her sick grandmother. \\
 Beat 21: Grandmother asks Red to download and install a specific software on her computer. \\
 Beat 3: Red, wearing her red hoodie, ventures out with her electric scooter. \\
 Beat 4: En route, Red encounters a stranger, a cunning hacker, who learns about her mission. \\
 Beat 22: The hacker sends Red a malicious software disguised as the one requested by her grandmother. \\
 Beat 23: Red, noticing the odd behavior of the software, realizes the trick. \\
Beat 24: Red, with the help of her tech community, removes the malicious software and exposes the hacker. \\
 Beat 25: Red installs the correct software on her grandmother's computer.

Storyline 6: \\
 Beat 1: Red, a tech-savvy girl living in a smart city, receives a call from her sick grandmother. \\
 Beat 26: Grandmother asks Red to bring her some digital books from the e-library. \\
 Beat 3: Red, wearing her red hoodie, ventures out with her electric scooter. \\
 Beat 4: En route, Red encounters a stranger, a cunning hacker, who learns about her mission. \\
 Beat 27: The hacker tries to gain access to Red's e-library account. \\
Beat 28: Red, noticing the login attempts, secures her account and reports the hacker. \\
 Beat 29: Red downloads the digital books and delivers them to her grandmother.

Storyline 7: \\
 Beat 15: Red, a social media influencer, plans a live stream to visit her grandmother. \\
 Beat 21: Grandmother asks Red to download and install a specific software on her computer. \\
 Beat 3: Red, wearing her red hoodie, ventures out with her electric scooter. \\
 Beat 16: Red's live stream attracts the attention of a cyber-stalker. \\
 Beat 30: The stalker sends Red a dangerous link pretending to be a fan. \\
 Beat 31: Red, being tech-savvy, recognizes the dangerous link and alerts her followers. \\
 Beat 32: Red's followers report the stalker leading to his arrest. \\
 Beat 25: Red installs the correct software on her grandmother's computer.

Storyline 8: \\
 Beat 15: Red, a social media influencer, plans a live stream to visit her grandmother. \\
 Beat 26: Grandmother asks Red to bring her some digital books from the e-library. \\
 Beat 3: Red, wearing her red hoodie, ventures out with her electric scooter. \\
 Beat 16: Red's live stream attracts the attention of a cyber-stalker. \\
 Beat 33: The stalker tries to disrupt Red's live stream by spreading false rumors. \\
 Beat 34: Red, noticing the disruption, uses her influence to debunk the rumors. \\
 Beat 35: Red's followers, united, report the stalker leading to his arrest. \\
Beat 29: Red downloads the digital books and delivers them to her grandmother. \\ \\
START\_1: Points to Beat 1 \\
START\_2: Points to Beat 15 \\
END\_1: Points from Beat 9\\
 END\_2: Points from Beat 14\\
 END\_3: Points from Beat 25\\
 END\_4: Points from Beat 29

Beats:\\
Beat 1: Red, a tech-savvy girl living in a smart city, receives a call from her sick grandmother.\\
 Beat 2: Grandmother requests Red to bring her some medicines from the nearby pharmacy.\\
 Beat 3: Red, wearing her red hoodie, ventures out with her electric scooter.\\
 Beat 4: En route, Red encounters a stranger, a cunning hacker, who learns about her mission.\\
 Beat 5: The hacker manipulates the city's GPS system to mislead Red.\\
 Beat 6: Misled, Red ends up in an abandoned factory.\\
 Beat 7: Realizing the trick, Red uses her tech skills to trace the hacker's location.\\
 Beat 8: Red exposes the hacker to the city's cyber police and continues her journey to her grandmother's house.\\
 Beat 9: Red delivers the medicines and they have a virtual family gathering via video call.\\
 Beat 10: Grandmother asks Red to bring her a special gadget from the tech mall.\\
 Beat 11: The hacker hacks into Red's smartwatch, stealing her personal data.\\
 Beat 12: Red notices suspicious activity on her smartwatch and seeks help from her friend, a cybersecurity expert.\\
 Beat 13: Together, they trace the hacker and retrieve Red's data.\\
 Beat 14: Red buys the gadget and delivers it to her grandmother.\\
 Beat 15: Red, a social media influencer, plans a live stream to visit her grandmother.\\
 Beat 16: Red's live stream attracts the attention of a cyber-stalker.\\
 Beat 17: The stalker tries to find Red's location using the live stream data.\\
 Beat 18: The stalker tries to manipulate Red's followers against her.\\
 Beat 19: Red, noticing the unusual comments, uses her influence to expose the stalker's intentions.\\
 Beat 20: Red's followers, united, report the stalker leading to his arrest.\\
 Beat 21: Grandmother asks Red to download and install a specific software on her computer.\\
 Beat 22: The hacker sends Red a malicious software disguised as the one requested by her grandmother.\\
 Beat 23: Red, noticing the odd behavior of the software, realizes the trick.\\
 Beat 24: Red, with the help of her tech community, removes the malicious software and exposes the hacker.\\
 Beat 25: Red installs the correct software on her grandmother's computer.\\
Beat 26: Grandmother asks Red to bring her some digital books from the e-library.\\
 Beat 27: The hacker tries to gain access to Red's e-library account.\\
 Beat 28: Red, noticing the login attempts, secures her account and reports the hacker.\\
 Beat 29: Red downloads the digital books and delivers them to her grandmother.\\
 Beat 30: The stalker sends Red a dangerous link pretending to be a fan.\\
 Beat 31: Red, being tech-savvy, recognizes the dangerous link and alerts her followers.\\
 Beat 32: Red's followers report the stalker leading to his arrest.\\
Beat 33: The stalker tries to disrupt Red's live stream by spreading false rumors.\\
Beat 34: Red, noticing the disruption, uses her influence to debunk the rumors.\\
Beat 35: Red's followers, united, report the stalker leading to his arrest.

Common intermediate Beats: Beat 3, Beat 4, Beat 16 \\
    Storylines (8)\\
Storyline 1: START\_1, 1, 2, 3, 4, 5, 6, 7, 8, 9, END\_1\\
 Storyline 2: START\_1, 1, 10, 3, 4, 11, 12, 13, 14, END\_2\\
 Storyline 3: START\_2, 15, 2, 3, 16, 17, 7, 8, 9, END\_1\\
 Storyline 4: START\_2, 15, 10, 3, 16, 18, 19, 20, 14, END\_2\\
 Storyline 5: START\_1, 1, 21, 3, 4, 22, 23, 24, 25, END\_3\\
 Storyline 6: START\_1, 1, 26, 3, 4, 27, 28, 29, END\_4\\
  Storyline 7: START\_2, 15, 21, 3, 16, 30, 31, 32, 25, END\_3\\
 Storyline 8: START\_2, 15, 26, 3, 16, 33, 34, 35, 29, END\_4

\subsection{Prompt for generating graph}

Here we share the exact prompt used to generate the graph structure given the storylines (Section \ref{sec:generating-visualization})

INSTRUCTION: Given this narrative game draft [include the storylines i.e. the exact output generated by the model on the previous prompt], your task is to structure this input as nodes and edges objects striclty following the format described below.

Guideline 1: For example, take a story draft structured as follows:  \\
 Story: Little Red Riding Hood,\\
 Starts: 1,\\
 Endings: 1,\\
 Storylines: 8,\\
 Setting: Minecraft\\\\
 START\_1: (This is a dummy node. No description for it. It will always point to the beginning beat of the respective storyline)\\
 END\_1: (This is a dummy node. No description for it. The final node of the respective storyline will point to it.)

Beats:\\
 Beat\_1: Little Red Riding Hood, a Minecraft character, is given a task by her mother to deliver a basket of food to her grandmother's house.\\
 Beat\_2: Little Red Riding Hood ventures through a dense forest biome, collecting materials for her journey.\\
 Beat\_3: She encounters a friendly Minecraft villager who warns her about the dangerous wolves in the forest.\\
 Beat\_4: Little Red Riding Hood is distracted by a beautiful flower biome and strays off the path.\\
 Beat\_5: She encounters a wolf (a Minecraft mob), who tricks her into revealing her grandmother's location.\\
 Beat\_6: The wolf races ahead and locks her grandmother in a Minecraft dungeon.\\
 Beat\_7: Little Red Riding Hood arrives at her grandmother's house and realizes something is wrong.\\
 Beat\_8: She bravely confronts the wolf and rescues her grandmother by using her Minecraft tools.

Common intermediate beats: Beat\_3, Beat\_5\\
Storylines (8):\\
 Storyline 1: START\_1, Beat\_1, Beat\_2, Beat\_3, Beat\_5, Beat\_7, Beat\_8, END\_1\\
 Storyline 2: START\_1, Beat\_1, Beat\_2, Beat\_3, Beat\_4, Beat\_5, Beat\_8, END\_1\\
 Storyline 3: START\_1, Beat\_1, Beat\_2, Beat\_3, Beat\_5, Beat\_6, Beat\_7, Beat\_8, END\_1\\
 Storyline 4: START\_1, Beat\_1, Beat\_2, Beat\_4, Beat\_3, Beat\_5, Beat\_7, Beat\_8, END\_1\\
 Storyline 5: START\_1, Beat\_1, Beat\_3, Beat\_2, Beat\_4, Beat\_5, Beat\_8, END\_1\\
 Storyline 6: START\_1, Beat\_1, Beat\_3, Beat\_2, Beat\_5, Beat\_6, Beat\_7, Beat\_8, END\_1\\
 Storyline 7: START\_1, Beat\_1, Beat\_3, Beat\_2, Beat\_5, Beat\_7, Beat\_8, END\_1\\
 Storyline 8: START\_1, Beat\_1, Beat\_3, Beat\_5, Beat\_2, Beat\_4, Beat\_7, Beat\_8, END\_1

Guideline 2: Now, consider the next convention for nodes and edges objects from a network representing the given storylines. \\
These objects are meant as input data to a Javascript D3JS browser application for visualization. Bear in mind START and END nodes are always in the end of each object.\\
NODES:\\
\{\\
 "Beat\_1": [["None", 1, "Little Red Riding Hood, a Minecraft character, is given a task by her mother to deliver a basket of food to her grandmother's house.", "1"]],\\
 "Beat\_2": [["None", 2, "Little Red Riding Hood ventures through a dense forest biome, collecting materials for her journey.", "1"]],\\
  "Beat\_3": [["None", 3, "She encounters a friendly Minecraft villager who warns her about the dangerous wolves in the forest.", "1"]],\\
 "Beat\_4": [["None", 4, "Little Red Riding Hood is distracted by a beautiful flower biome and strays off the path.", "1"]],\\
 "Beat\_5": [["None", 5, "She encounters a wolf (a Minecraft mob), who tricks her into revealing her grandmother's location.", "1"]],\\
  "Beat\_6": [["None", 6, "The wolf races ahead and locks her grandmother in a Minecraft dungeon.", "1"]],\\
 "Beat\_7": [["None", 7, "Little Red Riding Hood arrives at her grandmother's house and realizes something is wrong.", "1"]],\\
 "Beat\_8": [["None", 8, "She bravely confronts the wolf and rescues her grandmother by using her Minecraft tools.", "1"]],\\
 "START\_1": [["None", null, null, null]],\\
 "END\_1": [["None", null, null, null]]\\
 \} \\
EDGES:\\
{\\
 "Beat\_1": {"None": [[["START\_1", "Beat\_1"]], [["Beat\_1", "Beat\_2"], ["Beat\_1", "Beat\_3"]]]}, \\
 "Beat\_2": {"None": [[["Beat\_1", "Beat\_2"]], [["Beat\_2", "Beat\_3"], ["Beat\_2", "Beat\_4"]]]},\\
 "Beat\_3": {"None": [[["Beat\_1", "Beat\_3"],["Beat\_2", "Beat\_3"]], [["Beat\_3", "Beat\_4"], ["Beat\_3", "Beat\_5"]]]}, \\
 "Beat\_4": {"None": [[["Beat\_2", "Beat\_4"], ["Beat\_3", "Beat\_4"]], [["Beat\_4", "Beat\_5"]]]}, \\
 "Beat\_5": {"None": [[["Beat\_3", "Beat\_5"], ["Beat\_4", "Beat\_5"]], [["Beat\_5", "Beat\_6"], ["Beat\_5", "Beat\_7"]]]},\\
 "Beat\_6": {"None": [[["Beat\_5", "Beat\_6"]], [["Beat\_6", "Beat\_7"]]]},\\
 "Beat\_7": {"None": [[["Beat\_5", "Beat\_7"], ["Beat\_6", "Beat\_7"]], [["Beat\_7", "Beat\_8"]]]}, \\
 "Beat\_8": {"None": [[["Beat\_7", "Beat\_8"]], [["Beat\_8", "END\_1"]]]},\\
 "START\_1": {"None": [[], [["START\_1", "Beat\_1"]]]},\\
 "END\_1": {"None": [[["Beat\_8", "END\_1"]],[]]}\\
\}\\
More guidelines:
\begin{enumerate}
    \setlength\itemsep{0.1em}
\item Notice the meaning of elements in the nodes representation: {node\_id: [[game\_state, nr\_beat, beat, pathway]]}, where:
node\_id is a string with the label "Beat\_" and a number to identify a node, game\_state is the game state, nr\_beat is the number of the respective beat, beat is a string describing respective beat, 
pathway is a string with an integer label to identify the path in the graph corresponding to a quest or storyline.
\item Each node must correspond to one and only one beat, so that the number of nodes and beats are the same in the end.
\item Make sure to create a node for every beat. No beat should be left without a node.
\item Don't create nodes semantically equal. Each node has a unique and distinct beat associated to it in terms of semantic.            
\item For every beginning beat, create an associated dummy START node (e.g. START\_1, START\_2, ...) and connect the latter to the former.
\item For every ending beat, create an associated dummy END node (e.g. END\_1, END\_2, ...) and connect the former to the latter.
\item Make sure to create an edge between each pair of adjacent nodes in the given sequences for the storylines. Make sure you don't miss out any edge.
\item Every node must be connected to the graph.
\item START nodes must be at the end of the NODES and EDGES objects. START nodes are prohibited in the beginning of any objects. NEVER EVER put START and END nodes in the beginnig of any object.
\item END nodes must be at the end of the NODES and EDGES objects. END nodes are prohibited in the beginning of any object. NEVER EVER put START and END nodes in the beginnig of any object.
\item Make sure that every node in the NODES object also appears in the EDGES object and vice-versa.
\item Color the nodes pertaining to a same storyline with the very same color, that is, assigning a same integer value starting from 1 to the correspoding pathline property of the node.
\end{enumerate}

\end{document}